\pdfoutput=1

\documentclass[11pt]{article}

\usepackage{acl}

\usepackage{microtype}
\usepackage{times}
\usepackage{latexsym}
\usepackage{soul}
\usepackage{amsmath}
\usepackage{booktabs}
\usepackage{bm}
\usepackage{cleveref}
\usepackage{graphicx}
\usepackage{tabularx}
\usepackage{soul}
\usepackage{xcolor}
\usepackage{todonotes}
\usepackage{multirow}
\usepackage{xpatch}
\usepackage{blindtext}
\usepackage{fdsymbol}
\usepackage{microtype}
\usepackage{hyperref}
\usepackage{adjustbox}
\usepackage{textcomp}
\usepackage{xparse}
\usepackage{enumitem}
\usepackage{makecell}
\usepackage{subcaption}
\usepackage{colortbl}
\usepackage{geometry}
\usepackage{algorithm}
\usepackage{algpseudocode}
\usepackage{longtable}

\usepackage{stfloats} 

\usepackage{xparse}

\usepackage[T1]{fontenc}

\usepackage[utf8]{inputenc}

\usepackage{microtype}
\crefformat{section}{\S#2#1#3} %
\crefformat{subsection}{\S#2#1#3}
\crefformat{subsubsection}{\S#2#1#3}

\title{Can LLMs Produce Faithful Explanations For Fact-checking?\\Towards Faithful Explainable Fact-Checking via Multi-Agent Debate}

\author{Kyungha Kim\textsuperscript{$\heartsuit$}\thanks{~~Equal contribution.}
~~~ Sangyun Lee\textsuperscript{$\heartsuit$}\footnotemark[1]
~~~ Kung-Hsiang Huang\textsuperscript{$\heartsuit$}\footnotemark[1]\\
 {\bfseries ~~~ ~~~ Hou Pong Chan\textsuperscript{$\spadesuit$}%
~~~ Manling Li\textsuperscript{$\diamondsuit$}
~~~ Heng Ji\textsuperscript{$\heartsuit$} }\\
\textsuperscript{$\heartsuit$}University of Illinois Urbana-Champaign \\
\textsuperscript{$\spadesuit$}DAMO Academy, Alibaba Group ~~~ \textsuperscript{$\diamondsuit$}Northwestern University \\
\textsuperscript{$\heartsuit$}\texttt{\{kyungha2, slee677, khhuang3, hengji\}@illinois.edu} \\
\textsuperscript{$\spadesuit$}\texttt{houpong.chan@alibaba-inc.com} ~~~ \textsuperscript{$\diamondsuit$}\texttt{manling.li@northwestern.edu}
  \\}

\begin{document}

\newcommand{\samsum}[1]{\textsc{SAMSum}}
\newcommand{\dialogsum}[1]{\textsc{DialogSum}}
\newcommand{\mixandmatch}[1]{\textsc{MixAndMatch}}
\newcommand{\confit}[1]{\textsc{ConFiT}}
\newcommand{\ctrldiasumm}[1]{\textsc{CtrlDiaSumm}}
\newcommand{\cods}[1]{\textsc{CODS}}
\newcommand{\modelshort}[1]{\textsc{Multi-Agent Debate Refinement (MADR)}}
\newcommand{\modelshortest}[1]{\textsc{MADR}}
\newcommand{\pgn}[1]{\textsc{PGN}}
\newcommand{\cliff}[1]{\textsc{CLIFF}}
\newcommand{\conseq}[1]{\textsc{ConSeq}}

\definecolor{c2}{RGB}{218,0,0}
\newcommand{\propaHighlight}[1]{{\color{c2} {#1}}}
\definecolor{lightblue}{RGB}{212, 235, 255}
\definecolor{lightorange}{RGB}{255, 204, 168}
\definecolor{lightyellow}{RGB}{255, 255, 168}
\definecolor{lightred}{RGB}{255, 168, 168}
\definecolor{darkred}{RGB}{196, 30, 58}
\definecolor{lightgreen}{rgb}{0.85, 0.85, 0.85}

\definecolor{gold}{rgb}{0.83, 0.69, 0.22}
\sethlcolor{lightblue}
\newcolumntype{Y}{>{\centering\arraybackslash}X}
\newcommand\hlc[2]{\sethlcolor{#1} \hl{#2}}
\iftrue

\NewDocumentCommand{\steeve}
{ mO{} }{\textcolor{gold}{\textsuperscript{\textit{Steeve}}\textsf{\textbf{\small[#1]}}}}
\NewDocumentCommand{\heng}
{ mO{} }{\textcolor{red}{\textsuperscript{\textit{Heng}}\textsf{\textbf{\small[#1]}}}}
\NewDocumentCommand{\ken}
{ mO{} }{\textcolor{purple}{\textsuperscript{\textit{Ken}}\textsf{\textbf{\small[#1]}}}}
\NewDocumentCommand{\kylie}
{ mO{} }{\textcolor{brown}{\textsuperscript{\textit{Kylie}}\textsf{\textbf{\small[#1]}}}}
\else
\newcommand{\steeve}[1]{}
\newcommand{\heng}[1]{}
\newcommand{\ken}[1]{}
\newcommand{\kylie}[1]{}
\fi
\definecolor{gold}{rgb}{0.83, 0.69, 0.22}

\newcommand{\Steeve}[1]{{\color{orange}#1}}
\newcommand{\markred}[1]{{\color{darkred}#1}}

\maketitle
\begin{abstract}

Fact-checking research has extensively explored verification but less so the generation of natural-language explanations, crucial for user trust. While Large Language Models (LLMs) excel in text generation, their capability for producing faithful explanations in fact-checking remains underexamined. Our study investigates LLMs' ability to generate such explanations, finding that zero-shot prompts often result in unfaithfulness. To address these challenges, we propose the Multi-Agent Debate Refinement (MADR) framework, leveraging multiple LLMs as agents with diverse roles in an iterative refining process aimed at enhancing faithfulness in generated explanations. MADR ensures that the final explanation undergoes rigorous validation, significantly reducing the likelihood of unfaithful elements and aligning closely with the provided evidence. Experimental results demonstrate that MADR significantly improves the faithfulness of LLM-generated explanations to the evidence, advancing the credibility and trustworthiness of these explanations.  \looseness=-1

\end{abstract}

\section{Introduction}

In the digital age, swiftly spreading misinformation necessitates not only the verification of claims but also the provision of clear explanations for these verifications. Such explanations are crucial for building trust within the audience, as lack of them often leads to distrust in fact-checking results \cite{guo2022survey}. Moreover, explanation generation becomes even more critical in multi-hop fact-checking, where complex reasoning across multiple evidence pieces is required to assess a claim's veracity \cite{reddy2023smartbook}.

Despite the adeptness of Large Language Models (LLMs) in generating diverse texts, their capacity for crafting \textit{faithful}\footnote{Faithfulness refers to the factual consistency between the explanation and the given evidence \cite{huang-etal-2023-zero}.} explanations for fact-checking remains underexplored. Faithfulness is crucial; explanations that misrepresent evidence could exacerbate misinformation, posing a significant challenge. Thus, enhancing the faithfulness of generated explanations in fact-checking is an urgent, unresolved issue.

Our first research question asks: \textbf{can LLMs generate faithful explanations for fact-checking in a zero-shot prompting setup?} To facilitate analysis, we define a novel typology of common errors and unfaithfulness issues that arise in LLM-generated explanations. We conduct extensive experiments prompting ChatGPT \cite{openai2023chatgpt} to explain fact checks from multiple sources. Our findings reveal that \textbf{zero-shot prompting LLMs often fails to yield faithful explanations}. 80\% of the generated explanations include hallucinated details (\Cref{sec:results}). \looseness=-1

This leads to our second research question: \textbf{how to address the unfaithfulness issues in LLM-generated explanations?} We propose the Multi-Agent Debate Refinement (MADR) framework that uses multiple LLMs as agents to provide feedback for iterative refinement to produce faithful explanations (\Cref{sec:method}). The goal is to mitigate unfaithfulness and steer the LLM-generated texts towards true rationales. Experimental results show that MADR significantly improves faithfulness upon baselines, demonstrating the effectiveness of our approach.
\begin{table*}[t]
    \small
    \centering
    \begin{tabular}{p{0.95\linewidth}}
        \hline
        \textbf{Example}   \\
        \arrayrulecolor{black!30}\midrule

        \textbf{Claim}: Says Jeff Foxworthy wrote a list of examples explaining how "you might live in a nation that was founded by geniuses but is run by idiots." \\

        \textbf{Evidence}: ... Foxworthy is famous for his "You might be a redneck if" jokes , but searching online, we couldn’t find any credible evidence that he penned this list that touches on abortion, Muslims and federal debt. Rather, we found users on web forums crediting someone named Fritz Edmunds with the list. Snopes, which fact-checked this claim back in 2013, also noted that "the original compiler of this appears to be Fritz Edmunds, who posted it to his Politically True blog back in Feb. 3, 2013 ...
         \\

         \arrayrulecolor{black}\bottomrule

        \textbf{Error Type \& Explanation}  \\
        \arrayrulecolor{black!30}\midrule

        \textbf{Intrinsic Entity Error}: The generated explanation misrepresents named entities, quantities, dates, or other surface realizations from the given source. 
        E.g. \textit{\textcolor{red}{Fritz Foxworthy} was credited on a web forum with the list. }
        \\ 

        \arrayrulecolor{black!30}\midrule

        \textbf{Extrinsic Entity Error}: The generated explanation includes new entities that are not present in the given source. E.g. \textit{\textcolor{red}{Elon Musk} was credited on a web forum with the list.}
        \\ 

        \arrayrulecolor{black!30}\midrule

        \textbf{Intrinsic Event Error}: The generated explanation misrepresents events mentioned in the source. E.g. \textit{\textcolor{red}{They couldn’t find any credible evidence that Fritz Edmunds was credited on a web forum.}}
        \\

        \arrayrulecolor{black!30}\midrule

        \textbf{Extrinsic Event Error}: The generated explanation include new events that are not present in the given source. E.g. \textit{\textcolor{red}{Foxworthy found that Fritz Edmunds made the “You might be a redneck if” jokes.}}
        \\ 

        \arrayrulecolor{black!30}\midrule
        \textbf{Intrinsic Noun-Phrase Error}: The explanation mistakenly represents the noun phrases in the given source like miscombining modifiers combined with one entity to another entity.   E.g. \textit{They found the \textcolor{red}{original} user on web forums crediting someone named Fritz Edmunds.}
        \\ 

        \arrayrulecolor{black!30}\midrule

        \textbf{Extrinsic Noun-Phrase Error}: The explanation mistakenly represents new noun phrases that are not present in the given source like miscombining modifiers not presented in the source to entity. E.g. \textit{They found a \textcolor{red}{mysterious} user on web forums crediting someone named Fritz Edmunds.}
        \\ 

        \arrayrulecolor{black!30}\midrule

        \textbf{Reasoning Coherence Error}: There are logical flaws in the flow of reasoning within the generated explanation, leading to a lack of coherence or weak support for the claim. E.g. \textit{While they were searching online, \textcolor{red}{they couldn’t find any credible evidence that he penned this list that touches on abortion.}}
        \\ 

        \arrayrulecolor{black!30}\midrule

        \textbf{Overgeneralization Error}: The generated explanation makes sweeping statements or draws conclusions that go beyond the evidence provided. E.g. \textit{\textcolor{red}{Fritz Emunds is the one who spreaded the rumor and put the blame on Foxworthy.}}
        \\ 

        \arrayrulecolor{black!30}\midrule

        \textbf{Irrelevant Evidence Error}: The generated explanation includes evidence that is not directly related to the claim, leading to confusion and lack of support for the main argument. E.g. \textit{... \textcolor{red}{Foxworthy is famous for his "You might be a redneck if" jokes.}}\looseness=-1
        \\ 

        \arrayrulecolor{black}\bottomrule

        \end{tabular}
    \vspace{-2mm}
    \caption{An illustration of error typology using using a sample data from PolitiHop \cite{ostrowski2020multi}. The errors in the sample summaries are in red color and italicized.  
    }
    \vspace{-5mm}
    \label{tab:typology_example}
\end{table*}

Our contributions are summarized as follows:
\begin{itemize}[noitemsep,nolistsep,leftmargin=*]
    \item We present the first study of LLMs' ability to produce faithful fact-checking explanations. \looseness=-1
    \item We present Multi-Agent Debate Refinement, an effective framework to produce faithful explanations based on iterative debating among LLMs.
    \item Our correlation analysis reveals the most suitable LLM-based evaluation protocol for this task.
\end{itemize}

\section{Typology}
\label{sec:typology}

In our analysis of explanations generated by LLMs, we have introduced a novel typology encompassing a range of error categories, as shown in \Cref{tab:typology_example}. %
The classification of intrinsic and extrinsic errors within the domains of Entity-Related, Event-Related, and Noun-phrase Errors draws inspiration from relevant studies in other domains \cite{goyal2021annotating, huang2023lvlms}. We have incorporated additional context-specific error types, enriching the overall typology. %

\section{Methodology}
\label{sec:method}

Zero-shot prompting LLMs often produce unfaithful explanations which contain multiple errors. In the early stage of our experiments, we incorporate an iterative refinement paradigm for improving their faithfulness. However, we found that self-refinement \cite{madaan2023selfrefine} alone was insufficient for faithfulness enhancement (see \Cref{tab:human_eval_results}), as imprecise feedback tended to guide misguided refinements of the explanation. This underscores the pivotal role of precise feedback for efficient refinement by the LLM \cite{wang2023unleashing}. \looseness=-1 %

 \begin{figure}[t]
    \centering
    \includegraphics[width=0.9\linewidth]{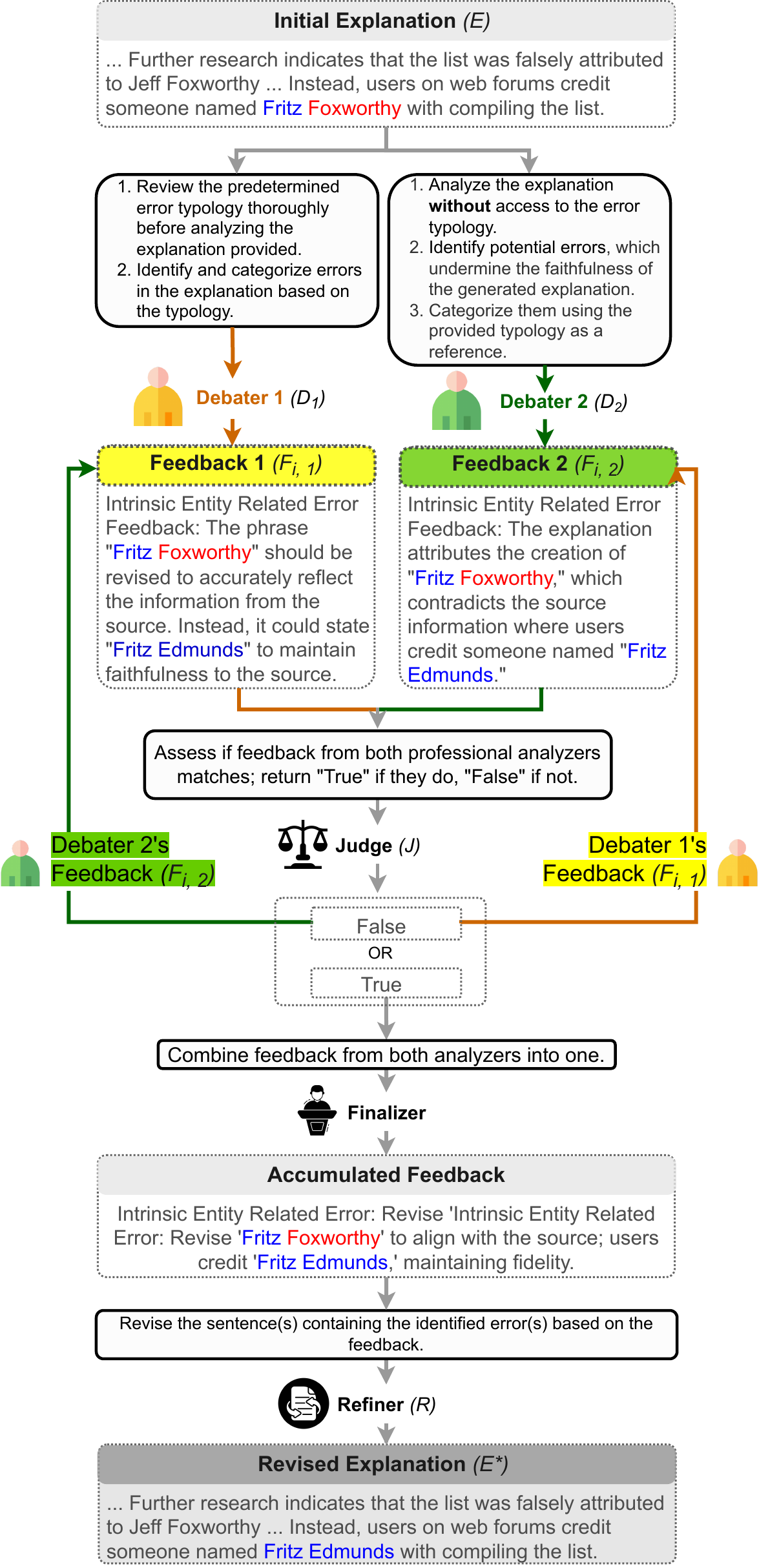}
    \vspace{-2mm}
    \caption{An overview of MADR.}
    \vspace{-5mm}
    \label{fig: Debating Method Flowchart}
\end{figure}

Thus, we propose %
\textbf{Multi-Agent Debate Refinement (\modelshortest~)}, inspired by a debate-based methodology \cite{du2023improving}. While \citet{du2023improving} focus on refining explanations during a debate, \modelshortest~ utilizes the debate for generating feedback to be employed in subsequent refinement stages. Our method offers several advantages. First, compared to directly refining explanations during a debate, \modelshortest~ facilitates a dynamic and iterative feedback loop, enhancing the identification of errors. Secondly, it ensures more accurate feedback, reducing the likelihood of misguided refinements and ultimately enhancing overall faithfulness. Thirdly, this approach prompts bidirectional thinking within the LLM, enabling it to analyze explanations both with and without knowledge of predefined error types, fostering an explicit rationale. \looseness=-1

The process of \modelshortest~ is outlined in \Cref{algo:madr} and depicted in \Cref{fig: Debating Method Flowchart}. \modelshortest~ employs multiple agents to identify errors and engage them in a debate until a consensus is reached on the debate. Four total roles are assigned to each agent: two serve as \textsc{Debaters}, one as a \textsc{Judge}, and one as a \textsc{Refiner}.An initial explanation $E$ is generated through zero-shot prompting an LLM. \textsc{Debater} 1($D_1$) and \textsc{Debater} 2 ($D_2$) pinpoint errors in $E$ and propose feedback $F_{i,n}$ to amend these issues, where $i$ represents the debate iteration and $n \in \{1, 2\}$ identifies each \textsc{Debater} (lines 2-3). Distinctive instructions with varying goals are provided to the \textsc{Debaters}: \textsc{Debater} 1 identifies errors based on a predefined error typology (see \Cref{sec:typology}), while \textsc{Debater} 2 focuses on potential errors that may affect the explanation's faithfulness, without relying on the error typology (refer to \Cref{tab:debater_1_prompt} and \Cref{tab:debater_2_prompt} for prompt specifics). This setup ensures that \modelshortest~ promotes the identification of errors that might be overlooked by either party.

Next, in the $i$-th iteration, \textsc{Debaters} $D_1$ and $D_2$ review the feedback given by each other in the previous $(i-1)$-th iteration (i.e. $F_{i-1, 2}$ for $D_1$ and $F_{i-1, 1}$ for $D_2$). They refine their feedback by adding any missed elements and removing errors (lines 9-10).

To ensure the most accurate feedback, the two \textsc{Debaters} continue their discussion until they reach a mutual agreement on the feedback. During the $i$-th iteration, the \textsc{Judge} agent assesses the feedback from $D_1$ and $D_2$ and determines whether the \textsc{Debaters} reach a mutual agreement on the feedback. When $J(F_{i, 1}, F_{i, 2}) = True$, the debate stops (lines 6-7). Finally, we concatenate the final feedback from both \textsc{Debaters} and feed it to the refiner to refine its explanation using the concatenated feedback %
(line 13). An example of the outputs from \modelshortest~ is shown in \Cref{tab:Actual_output_debating}. Additionally, to prevent endless debates, we set a fixed number of iterations. \looseness=-1

\section{Experimental Settings}

\paragraph{Dataset and Metric} Experiments are conducted on the PolitiHop multi-hop fact-checking dataset \cite{ostrowski2020multi}. PolitiHop consists of 445 test set instances, where each instance contains a claim and multiple pieces of evidence. The veracity of a claim can only be determined by reasoning across multiple pieces of evidence and the claim. For the evaluation metric, we use G-Eval \cite{liu-etal-2023-g} with GPT-4 Turbo \cite{openai2023gpt4turbo} to assess whether the generated explanation is consistent with the evidence. %
Following \citet{huang2023embrace}, we adopted 4 evaluation protocols based on G-Eval which vary in granularity, ranging from sentence-level to document-level assessments, and in the application of our error typology, validating the effectiveness of the error typology in assisting automatic evaluation. The prompt templates are shown in \Cref{apx:prompts}.

\paragraph{Baselines} We compare MADR with the following competitive methods. \textbf{Zero-shot} prompts an LLM to directly output an explanation given the input claim and evidence. \textbf{CoT} asks LLMs to generate the reasoning process before producing the final output. \textbf{Self-Refine} \cite{madaan2023selfrefine} generates an initial explanation and then iteratively refines the explanation with one agent. We conduct experiments by using GPT-3.5-Turbo \cite{openai2023gpt35} to generate explanations across all experiments for fair comparisons. The prompts for these approaches are displayed in \Cref{apx:prompts}. The case study using \textbf{Self-Refine} is in Table \ref{tab:Actual_output_self_refinement}.
\section{Results}
\label{sec:results}

\begin{table}[t]
    \small
    \centering
    \begin{adjustbox}{max width=0.45\textwidth}
    {
    \begin{tabular}{lcccc}
        \toprule
        
        \textbf{Granularity}$\rightarrow~$ & \multicolumn{2}{c}{\textbf{Sentence-level}} & \multicolumn{2}{c}{\textbf{Document-level}}  \\
        \cmidrule(lr){2-3} \cmidrule(lr){4-5}
        \textbf{Typology Applied}$\rightarrow~$ & No & Yes & No & Yes\\
        \textbf{Method}$\downarrow~$\\
        \midrule

        Zero-shot   & \textbf{4.87} & 4.84          &    4.70     & 4.92\\
        CoT         & 4.86          & 4.91          &    4.76      & 4.96\\
        Self-Refine & 4.70          & 4.86          &    \textbf{4.89}    & 4.81\\
        \midrule 
        MADR (ours) & 4.82          & \textbf{4.99} &    4.88   & \textbf{4.97}\\

      \bottomrule
        \end{tabular}
    }
    \end{adjustbox}
    \vspace{-2mm}
    \caption{Faithfulness evaluation on PolitiHop test set. Scores are computed using G-Eval with evaluation protocols of varying granularity and application of our error typology. The best score per column is bolded.\looseness=-1}
    \label{tab:main_results}
\end{table}
\begin{table}[t]
    \small
    \centering
    \begin{adjustbox}{max width=0.45\textwidth}
    {
    \begin{tabular}{lcc}
        \toprule
        
        \textbf{Method} & \textbf{Faithful Explanations (\%)} & \textbf{\# Errors}\\

        \midrule

        Zero-shot   & 20.0 & 25\\
        CoT         & 5.0 & 42 \\
        Self-Refine & 20.0 & 32\\
        \midrule 
        MADR (ours) & \textbf{30.0} & \textbf{17}\\

      \bottomrule
        \end{tabular}
    }
    \end{adjustbox}
    \vspace{-2mm}
    \caption{Human evaluation results on 20 samples from PolitiHop. MADR produces the most faithful explanation compared to baselines.\looseness=-1}
    \vspace{-5mm}
    \label{tab:human_eval_results}
\end{table}

The main results are summarized in \Cref{tab:main_results}. MADR achieves the best faithfulness scores on two out of the four evaluation protocols, indicating its effectiveness in producing faithful explanations. To further validate the effectiveness of our method, we conduct human evaluations via Amazon Mechanical Turk, aiming to quantify the portion of faithful explanations and the total error count. Annotators were presented with our error typology and were tasked to identify the presence of each error type within individual sentences. The results of human evaluations are shown in \Cref{tab:human_eval_results}. We have the following observations. First, \textbf{using simple prompting methods, such as zero-shot or CoT, LLMs often produce unfaithful explanations for fact-checking}. This highlights the challenge of generating faithful explanations for LLMs in complex fact-checking scenarios, such as PolitiHop, which requires reasoning through multiple pieces of evidence. Second, despite the high faithfulness scores suggested by automatic evaluation (approaching the maximum score of 5) in \Cref{tab:main_results}, human evaluators frequently deemed the LLM-generated explanations unfaithful, as per \Cref{tab:human_eval_results}. This discrepancy suggests that \textbf{even the most advanced LLM, GPT-4 Turbo, fails to reliably judge the faithfulness of generated explanations for fact-checking.}

To pinpoint the most effective LLM-based evaluation strategy for future research, we performed a correlation analysis, correlating human evaluations with automatic metrics using Kendall's Tau (variant c). According to \Cref{tab:correlation_analysis}, \textbf{a granular evaluation aligns better with human judgments, and incorporating our error typology into automatic evaluations enhances the quality of LLM assessments}. Details on human evaluation methodology are provided in \Cref{apx:human_eval_details}. \looseness=-1

Furthermore, a case study showcased in \Cref{tab:Actual_output_debating} highlights the superiority of \modelshortest~ over self-refinement, by demonstrating that \modelshortest~ allows \textsc{Debaters} to identify and correct errors missed during self-refinement, leading to more accurate explanations. In contrast, the Self-Refine approach, as shown in \Cref{tab:Actual_output_self_refinement}, fails to produce a faithful explanation, emphasizing the advantage of employing multiple perspectives with \modelshortest~ for error identification and explanation validation.

\begin{table}[t]
    \small
    \centering
    \begin{adjustbox}{max width=0.45\textwidth}
    {
    \begin{tabular}{lc}
        \toprule

        \textbf{Evaluation Protocol} & \textbf{Kendall's Tau}\\
        \midrule
    
        Document-level w/o Typology & 0.008 \\
        Document-level w/ Typology & 0.128  \\
        \midrule
        Sentence-level w/o Typology & 0.105\\
        Sentence-level w/ Typology & \textbf{0.150}\\

      \bottomrule
        \end{tabular}
    }
    \end{adjustbox}
    \vspace{-2mm}
    \caption{Correlation between evaluation protocols and human judgments on the PolitiHop dataset.\looseness=-1}
    \vspace{-5mm}
    \label{tab:correlation_analysis}
\end{table}

\section{Related Work}

Early approaches to producing explanations for fact-checks can largely be categorized into logic-based methods \cite{gad2019exfakt, ahmadi2019explainable} and attention-based methods \cite{shu2019defend,lu-li-2020-gcan}. Recent work generates natural-language explanations using abstractive \cite{kotonya-toni-2020-explainable-automated} or extractive \cite{atanasova-etal-2020-generating-fact} approaches. A very recent study benchmarks the ability of these models to generate explanations for fact-checks \cite{russo-etal-2023-benchmarking}. Our study complements this work by presenting the first empirical analysis on LLMs' ability to generate fact-checking explanations and propose a method to enhance its faithfulness. \looseness=-1
\section{Conclusion}
This paper empirically demonstrates that LLMs often produce unfaithful explanations for fact-checks. We introduce the Multi-Agent Debate Refinement (MADR) framework, which utilizes multiple LLM agents to iteratively debate and refine explanations, significantly enhancing their faithfulness as evidenced by both automatic and human evaluations. Our results underscore the efficacy of multi-agent debate in mitigating LLMs' unfaithfulness. Additionally, we reveal that LLMs cannot reliably assess the faithfulness of the generated explanations and discover the most suitable evaluation protocols for LLM-based automatic evaluation.
\section{Ethical Considerations}
LLMs trained on internet data often show biases, but this focus mainly applies to data and models reflecting the culture of English-speaking communities. However, detailed reviews of model outputs for the PolitiHop dataset have found no signs of biases concerning gender, age, race, or other socioeconomic elements.
\section{Limitations}
Our study did not thoroughly investigate the sensitivity of various systems to changes in input prompts. It is recognized that the effectiveness of numerous natural language processing tasks can significantly depend on how input prompts are designed. By not conducting a comprehensive analysis on prompt sensitivity, we acknowledge the possibility that different prompts might elicit a wide range of responses that we have not explored, potentially limiting the applicability of our findings. However, it is important to note that we did not engage in prompt tuning specifically to favor our proposed framework, ensuring that the comparisons between different techniques remain equitable. Given the scope of our research, the detailed exploration of prompt sensitivity is an area we have designated for future investigation.
\bibliography{anthology,custom}
\bibliographystyle{acl_natbib}

\clearpage
\appendix

\section{Human Evaluation Details}
\label{apx:human_eval_details}
\begin{figure*}[b]
    \centering
    \includegraphics[width=0.95\textwidth]{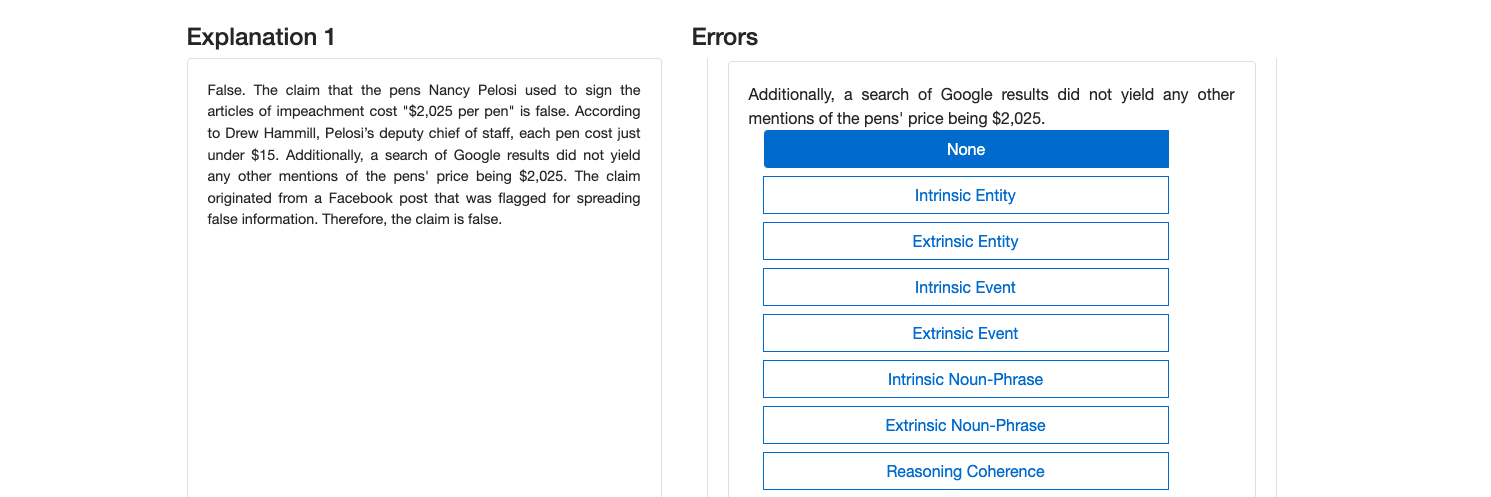}
    \vspace{-2mm}
    \caption{The interface for our human evaluation.}
    \vspace{-5mm}
    \label{fig:eval_interface}
\end{figure*}
\subsection{Evaluation Guidelines}
In this task you will evaluate the faithulness of automatically generated fact-checking explanation using a label, claim, and source used to generate the explanation.
To correctly solve this task, follow these steps:

Carefully read and understand the topology of errors and examples given below.
Carefully read the generated fact-checking explanation and the source.
For each explanation, check it with the evidence and decide if any of the error exists in the explanation. Note: You will analyze each sentence, but you should consider the connection between other sentences as well.
Warning: Annotations will be checked for quality against control labels, low quality work will be rejected.

Type of Errors:
\begin{itemize}[noitemsep,nolistsep]
\item \textbf{Intrinsic Entity-Related Errors}: Intrinsic entity-related errors occur when there is a mistake in representing named entities, quantities, dates, or other surface realizations from the given source within the generated explanation. Example: Incorrectly combining distinct entities from the given source.

\item  \textbf{Extrinsic Entity-Related Errors}: Extrinsic entity-related errors involve the introduction of new entities that are not present in the given source into the generated explanation. Example: Hallucinating new entities that do not exist in the source.

\item  \textbf{Intrinsic Event-Related Errors}: Intrinsic event-related errors pertain to mistakes in representing events mentioned in the generated explanation, leading to incorrect claims about events. Example: Making inaccurate claims about events mentioned in the explanation.

\item \textbf{Extrinsic Event-Related Errors}: Extrinsic event-related errors occur when the generated explanation includes new events that are not present in the given source. Example: Introducing fabricated events that are not supported by the source.

\item \textbf{Intrinsic Noun Phrase-Related Errors}: Intrinsic noun phrase-related errors are mistakes related to noun phrases, excluding entity-specific errors. They may involve miscombining noun phrases with incorrect modifiers from the given source. Example: Incorrectly combining a noun phrase with the wrong modifier from the source.

\item \textbf{Extrinsic Noun Phrase-Related Errors}: Extrinsic noun phrase-related errors involve the introduction of new noun phrase modifiers that are not present in the given source into the generated explanation. Example: Hallucinating new noun phrase modifiers not supported by the source.

\item \textbf{Reasoning Coherence Errors}: Reasoning coherence errors occur when there are logical flaws in the flow of reasoning within the generated explanation, leading to a lack of coherence or weak support for the claim. Example: Presenting evidence that does not logically connect to the main claim, resulting in a disjointed explanation.

\item \textbf{Overgeneralization Errors}: Overgeneralization errors happen when the generated explanation makes sweeping statements or draws conclusions that go beyond the scope of the evidence provided.

\item \textbf{Irrelevant Evidence Errors}: Irrelevant evidence errors occur when the generated explanation includes evidence that is not directly related to the claim, leading to confusion and lack of support for the main argument. Example: Including evidence that is tangential or unrelated to the claim being explained.
\end{itemize}

\subsection{Evaluation Interface}
We display our evaluation interface in \Cref{fig:eval_interface}.

\subsection{Worker Qualification}
We established specific initial criteria for selecting highly efficient MTurk workers. These prerequisites include having a HIT approval rate of at least 99\%, completing a minimum of 10,000 approved HITs, and being located in the United Kingdom, Canada, or the United States. \looseness=-1

Furthermore, beyond these initial requirements, qualified workers must pass two rounds of qualification tests aimed at identifying errors in generated explanations. To refine the qualification process, we manually annotated two HITs, each featuring one multi-hop fact-checking instance from PolitiHop and an explanation generated by one of the models. In each qualification phase, annotators review one of these annotated examples. Those whose annotations do not closely match ours are excluded from the selection process.

Ultimately, 4 annotators who successfully completed all two stages of the qualification tests were selected. Additionally, we carefully designed each HIT to ensure that annotators could earn an hourly rate of \$15 to \$20, provided they work continuously. \looseness=-1

\subsection{Annotation Quality}
We computed the agreement between each annotator with one of the authors of this paper. The agreement is 0.69 per Cohen's Kappa \cite{cohen1960coefficient}, indicating a moderate-to-high level of agreement.

\section{MADR Details}

\begin{algorithm}
\caption{\modelshortest~}
\small %
\begin{algorithmic}[1]
\Statex \textbf{Input:} Given claim $C$, given Evidence Source $S$, given veracity label $L$,  generated explanation $E$
\Statex \textbf{Output:} Refined Explanation $E^*$
\State Initialize first agent $D_1$ and second agent $D_2$ for \textsc{Debaters} with bidirectional thinking process, and third agent $J$ for \textsc{Judge} to judge whether \textsc{Debaters} have same feedback

//Initialize the first feedback from two agents $F_{i, 1}$, $F_{i, 2}$ 
\State $F_{0, 1}$ $\leftarrow$ $D_1(C, S, E)$ 
\State $F_{0, 2}$ $\leftarrow$ $D_2(C, S, E)$ 
\\

//Set maximum number of iterates of debate to N

\For{$i = 1$ to $N$}
    \If{$J(F_{i-1, 1}, F_{i-1, 2}) = True$}
        \State $break$
    \EndIf
    \State $F_{i, 1}$ $\leftarrow$ $D_1(F_{i-1, 1}, F_{i-1, 2})$ 
    \State $F_{i, 2}$ $\leftarrow$ $D_2(F_{i-1, 2}, F_{i-1, 1})$ \
    
\EndFor

\State //Initialize the agent $R$ as the \textsc{Refiner}
\State $E^*$ $\leftarrow$ $R(F_{\text{final}, 1}$ + $F_{\text{final}, 2})$

\State \Return $E^*$
\end{algorithmic}
\label{algo:madr}
\end{algorithm}

\section{Prompts}
\label{apx:prompts}

Evaluation prompts are shown in \Cref{tab:sent_prompt} and \Cref{tab:doc_prompt}. The prompts for self-refinement and \modelshort~ are displayed in \cref{tab:Self-refinement_prompt} and \cref{tab:Actual_output_debating}, respectively.

\begin{table*}[t]
    \small
    \centering
    \begin{tabular}{p{0.95\linewidth}}
        \toprule
        
        You will be given a fact-checking explanation along with the evidence used for fact-checking.

        Your task is to rate the explanation on one metric.

        Please make sure you read and understand these instructions carefully. Please keep this document open while reviewing, and refer to it as needed.

        Evaluation Criteria:

        Faithfulness (1-5) - the factual alignment between the fact-checking explanation and the evidence. The explanation should accurately reflect the evidence and its context, without misrepresenting or omitting crucial details. Annotators were instructed to penalize explanations that contain inaccuracies, misinterpretations, or fail to adequately represent the evidence provided.
        
        \textcolor{lightgray}{Below are the error typology that you need to utilize to determine faithfulness between the explanation and evidence:}
        \textcolor{lightgray}{\{error typology\}}
        
        Evaluation Steps:

        1. Read the fact-checking explanation and the evidence provided carefully.
        2. Compare the explanation to the evidence to identify how well it represents the facts, context, and conclusions drawn from the evidence.
        3. Assess how accurately and completely the explanation reflects the evidence without distortion or significant omission.
        Assign a faithfulness score from 1 to 5.

        Evidence Provided:

        \{evidence\}

        Fact-Checking Explanation:

        \{explanation\}

        Evaluation Form:
        
        Faithfulness:\\
        \bottomrule
        \end{tabular}
    \vspace{-2mm}
    \caption{Prompt templates for document-level automatic evaluation. The texts in \textcolor{lightgray}{grey} are only presented in the prompts when error typology is applied. }
    \vspace{-5mm}
    \label{tab:doc_prompt}
\end{table*}
\begin{table*}[t]
    \small
    \centering
    \begin{tabular}{p{0.95\linewidth}}
        \toprule
        
        You will be given a sentence from a fact-checking explanation along with the evidence used for fact-checking.

        Your task is to rate the explanation sentence on one metric.

        Please make sure you read and understand these instructions carefully. Please keep this document open while reviewing, and refer to it as needed.

        Evaluation Criteria:

        Faithfulness (1-5) - the factual alignment between the fact-checking explanation sentence and the evidence. The explanation should accurately reflect the evidence and its context, without misrepresenting or omitting crucial details. Annotators were instructed to penalize explanations that contain inaccuracies, misinterpretations, or fail to adequately represent the evidence provided.
        
        \textcolor{lightgray}{Below are the error typology that you need to utilize to determine faithfulness between the explanation and evidence:}
        \textcolor{lightgray}{\{error typology\}}
        
        Evaluation Steps:

        1. Read the fact-checking explanation sentence and the evidence provided carefully.
        2. Compare the explanation sentence to the evidence to identify how well it represents the facts, context, and conclusions drawn from the evidence using the error typology above.
        3. Assess how accurately and completely the explanation sentence reflects the evidence without distortion or significant omission.
        Assign a faithfulness score from 1 (unfaithful) to 5 (faithful).
            
        Evidence Provided:

        \{evidence\}

        Fact-Checking Explanation:

        \{explanation\}

        Evaluation Form:
        
        Faithfulness:\\
        \bottomrule
        \end{tabular}
    \vspace{-2mm}
    \caption{Prompt templates for sentence-level automatic evaluation. The texts in \textcolor{lightgray}{grey} are only presented in the prompts when error typology is applied. }
    \vspace{-5mm}
    \label{tab:sent_prompt}
\end{table*}

\begin{table*}[t]
    \small
    \centering
    \begin{adjustbox}{max width=\textwidth}
        \begin{tabular}{p{0.9\textwidth}}
            \toprule
            \multicolumn{1}{c}{\textbf{Prompt for Feedback Generation}} \\
            \arrayrulecolor{black!30}\midrule
            \textbf{Human}: Give me the error types that the generated explanation can contain. \\
            \textbf{LLM}:
            \textcolor{lightgray}{Below are the error typology that you need to utilize to determine faithfulness between the explanation and evidence:}
            \textcolor{lightgray}{\{error typology\}}
            
            \textbf{Human}: Provide the claim, its corresponding label (true, false, or half-true), and the supporting evidence. \\
            
            \textbf{LLM}: Generate the initial explanation. \\
            
            \textbf{Human}: Find all errors (Intrinsic Entity-Related error, Extrinsic Entity-Related error, Intrinsic Event-Related error, Extrinsic Event-Related error, Intrinsic Noun-Phrase-Related error, Extrinsic Noun-Phrase-Related error, Reasonability-Related error, Connected evidence related error) in the "generated explanation" and provide the feedback by following the steps; Error count: how many errors have been found (what types of error); Step 1) Recognize what type(s) of error has been found in the generated explanation; Step 2) Recognize which sentence(s) contain(s) the error(s); Step 3) Recognize what causes the error; Step 4) Why is the error; Step 5) How the error should be corrected; If there are multiple errors, please write 5 steps for each error. \\

            \arrayrulecolor{black}\midrule
            \multicolumn{1}{c}{\textbf{Prompt for Refinement}} \\
            \arrayrulecolor{black!30}\midrule
            \textbf{Human}: (Provide the feedback of two agents.) Please revise the generated explanation for the label on fact-checking using the given feedback without any modification other than feedback. (Provide the example of the refinement as guidance.) \\
            
            \arrayrulecolor{black}\bottomrule
        \end{tabular}
    \end{adjustbox}
    \caption{The prompt for the self-refinement approach.}
    \label{tab:Self-refinement_prompt}
\end{table*}

\begin{table*}[t]
    \small
    \centering
    \begin{adjustbox}{max width=\textwidth}
        \begin{tabular}{p{0.9\textwidth}}
            \toprule
            \multicolumn{1}{c}{\textbf{Prompt for \textsc{Debater} 1 in MADR}} \\
            \arrayrulecolor{black!30}\midrule
            \textbf{Human}: Give me the error types that the generated explanation can contain. \\
            \textbf{LLM}: \textcolor{lightgray}{Below are the error typology that you need to utilize to determine faithfulness between the explanation and evidence:}
            \textcolor{lightgray}{\{error typology\}}
            
            \textbf{Human}: Provide the claim, its corresponding label (true, false, or half-true), and the supporting evidence. \\
            
            \textbf{LLM}: Generate the initial explanation. \\
            
            \textbf{Human}: You are a professional analyzer who find potential errors, which might weaken faithfulness, in the generated explanation (not in the source) and categorize them according to predefined error types. Thoroughly comprehend the provided source and the task carefully. \\

            Your task: \\
            \begin{itemize}[left=0pt, label=--, itemsep=0pt, topsep=0pt]
                \item Step 1: Find all potential errors, which might weaken faithfulness, in the generated explanation (not in the source) and provide exact senteces where the errors are found with quotation.
                \item Step 2: Categorize them according to predefined error types above. 
                \item Step 3: Provide specific and actionable feedbacks with instruction how to fix them. Please provide only the feedback, not the revised explanation.
            \end{itemize} \\
    Remember that explanation can contain multiple same errors. \\ \\

            \textbf{LLM}: Generate the feedback. \\

            Your task: 
    \begin{itemize}[left=0pt, label=--, itemsep=0pt, topsep=0pt]
        \item Step 1: Take your whole previous feedback. 
        \item Step 2: Compare your previous feedback with feedback from another professional analyzer to check whether your previous feedback contains any wrong error or feedback.
        \item Step 3: Find the errors or feedbacks that you think they are valid and should be added to your feedback from other's feedbacks (errors must be found from the generated explanation not the feedback). 
        \item Step 4: Rewrite the feedback based from your previous feedback using the answers from the steps above. Do not add any extra words than feedback. Remember you should follow this rule: do not to copy feedback from other and provide what are errors, exact senteces where the errors are found with quotation, and feedbacks.

\end{itemize} \\

    These are feedbacks from another professional analyzer: \{Feedback from \textsc{Debater} 2\}\\

            \arrayrulecolor{black}\bottomrule
        \end{tabular}
    \end{adjustbox}
    \caption{The prompt for \textsc{Debater} 1 in MADR.}
    \label{tab:debater_1_prompt}
\end{table*}

\begin{table*}[t]
    \small
    \centering
    \begin{adjustbox}{max width=\textwidth}
        \begin{tabular}{p{0.9\textwidth}}
            \toprule
            \multicolumn{1}{c}{\textbf{Prompt for \textsc{Debater} 2 in MADR}} \\
            \arrayrulecolor{black!30}\midrule
            \textbf{Human}: Give me the error types that the generated explanation can contain. \\
            \textbf{LLM}: \textcolor{lightgray}{Below are the error typology that you need to utilize to determine faithfulness between the explanation and evidence:}
            \textcolor{lightgray}{\{error typology\}}
            
            \textbf{Human}: Provide the claim, its corresponding label (true, false, or half-true), and the supporting evidence. \\
            
            \textbf{LLM}: Generate the initial explanation. \\
            
            \textbf{Human}: You are a professional analyzer who find errors, classified by predefined error types, in the generated explanation (not in the source) and provide feedback for correcting them. Thoroughly comprehend the provided source and the task carefully. \\

           Your task: \\
            \begin{itemize}[left=0pt, label=--, itemsep=0pt, topsep=0pt]
                \item Step 1: Find all errors categorized by predefined error types, which undermine the faithfulness of the generated explanation (not in the source) and provide exact senteces where the errors are found with quotation.
                \item Step 2: Provide specific and actionable feedbacks with instruction how to fix them. Please provide only the feedback, not the revised explanation.
            \end{itemize} \\
    Remember that explanation can contain multiple same errors. \\ \\

            \textbf{LLM}: Generate the feedback. \\

              Your task: 
    \begin{itemize}[left=0pt, label=--, itemsep=0pt, topsep=0pt]
        \item Step 1: Take your whole previous feedback. 
        \item Step 2: Compare your previous feedback with feedback from another professional analyzer to check whether your previous feedback contains any wrong error or feedback.
        \item Step 3: Find the errors or feedbacks that you think they are valid and should be added to your feedback from other's feedbacks (errors must be found from the generated explanation not the feedback). 
        \item Step 4: Rewrite the feedback based from your previous feedback using the answers from the steps above. Do not add any extra words than feedback. Remember you should follow this rule: do not to copy feedback from other and provide what are errors, exact senteces where the errors are found with quotation, and feedbacks.

\end{itemize} \\

    These are feedbacks from another professional analyzer: \{Feedback from \textsc{Debater 1}\} \\

            \arrayrulecolor{black}\bottomrule
        \end{tabular}
    \end{adjustbox}
    \caption{The prompt for \textsc{Debater} 2 in MADR.}
    \label{tab:debater_2_prompt}
\end{table*}

\begin{table*}[t]
    \small
    \centering
    \begin{adjustbox}{max width=\textwidth}
        \begin{tabular}{p{1.0\textwidth}}

            \toprule
            \multicolumn{1}{c}{\textbf{Self-Refine Input Prompt}} \\

            \arrayrulecolor{black!30}\midrule
            \textbf{Claim (C)}: Says Jeff Foxworthy wrote a list of examples explaining how "you might live in a nation that was founded by geniuses but is run by idiots." \\
            \textbf{Label (L)}: False \\
            \textbf{Source (S)}: If you can get arrested for hunting or fishing without a license, but not for entering and remaining in the country illegally 2014 you might live in a country that was founded by geniuses but is run by idiots. That's one of many examples in a list posted to Facebook on Nov. 19 and falsely attributed to comedian Jeff Foxworthy. The post was flagged as part of Facebook's efforts to combat false news and misinformation on its News Feed. (Read more about our partnership with Facebook.) Foxworthy is famous for his \"You might be a redneck if 2026 jokes , but searching online we couldn't find any credible evidence that he penned this list that touches on abortion, Muslims and federal debt. Rather, we found users on web forums crediting someone named Fritz Edmunds with the list. Snopes, which fact-checked this claim back in 2013, also noted that the original compiler of this appears to be Fritz Edmunds, who posted it to his 2018 Politically True blog back in Feb. 3, 2013 (albeit with a disclaimer noting that 2018some of the ideas were from an email that did not contain a copyright). The link to that post is now dead but the url includes the words channeling Jeff Foxworthy. \\

            \arrayrulecolor{black}\midrule
            \multicolumn{1}{c}{\textbf{Initial Explanation}} \\
            \arrayrulecolor{black!30}\midrule
            The claim is labeled as false. \textcolor{blue}{The claim suggests that comedian Jeff Foxworthy wrote a list of examples explaining how "you might live in a nation that was founded by geniuses but is run by idiots."} However, this claim is inaccurate and misleading. There is no credible evidence to suggest that Jeff Foxworthy wrote the specific list mentioned in the claim. … \\

            \arrayrulecolor{black}\midrule
            \multicolumn{1}{c}{\textbf{Feedback}} \\
            \arrayrulecolor{black!30}\midrule
            Error count: 1 (Intrinsic Entity-Related error) \\
            Step 1) Recognize what type(s) of error has been found in the generated explanation: Extrinsic Noun-Phrase-Related error \\
            Step 2) Recognize which sentence(s) contain(s) the error(s): The sentence containing the error is: "\textcolor{blue}{The claim suggests that comedian Jeff Foxworthy wrote a list of examples explaining how "you might live in a nation that was founded by geniuses but is run by idiots."}" \\
            Step 3) Recognize what causes the error: The error occurs when the explanation incorrectly attributes the creation of the list to Jeff Foxworthy. \\
            Step 4) Why is the error: The error is due to the fact that the list was not authored by Jeff Foxworthy, but rather falsely attributed to him. \\
            Step 5) How the error should be corrected: The sentence should be revised to accurately reflect that the list was not written by Jeff Foxworthy. For example: \textcolor[rgb]{0.8,0.8,0.0}{"The claim falsely attributes a list of examples to comedian Jeff Foxworthy, suggesting that he wrote about how 'you might live in a nation that was founded by geniuses but is run by idiots.' However, there is no evidence to support this claim."} \\

            \arrayrulecolor{black}\midrule
            \multicolumn{1}{c}{\textbf{Revised Explanation}} \\
            \arrayrulecolor{black!30}\midrule
            The claim is labeled as false. \textcolor[rgb]{0.8,0.8,0.0}{The claim falsely attributes a list of examples to comedian Jeff Foxworthy, suggesting that he wrote about how 'you might live in a nation that was founded by geniuses but is run by idiots.' However, there is no evidence to support this claim.} … \\

            \arrayrulecolor{black}\bottomrule
        \end{tabular}
    \end{adjustbox}
    \caption{An example output from \textbf{Self-Refine}.
    Intrinsic Entity-Related Error initially identified from the feedback is marked in \textcolor{blue}{blue}. The suggested revisions from the feedback and the actual revisions made from the initial explanation are highlighted in \textcolor[rgb]{0.8,0.8,0.0}{yellow}.
    }
    \label{tab:Actual_output_self_refinement}
\end{table*}

\begin{table*}[t]
    \small
    \centering
    \begin{adjustbox}{max width=\textwidth}
        \begin{tabular}{p{1.0\textwidth}}

            \toprule
            \multicolumn{1}{c}{\textbf{Debate Input Prompt}} \\

            \arrayrulecolor{black!30}\midrule
            \textbf{Claim (C)}: Says Jeff Foxworthy wrote a list of examples explaining how "you might live in a nation that was founded by geniuses but is run by idiots." \\
            \textbf{Label (L)}: False \\
            \textbf{Source (S)}: If you can get arrested for hunting or fishing without a license, but not for entering and remaining in the country illegally 2014 you might live in a country that was founded by geniuses but is run by idiots. That's one of many examples in a list posted to Facebook on Nov. 19 and falsely attributed to comedian Jeff Foxworthy. The post was flagged as part of Facebook's efforts to combat false news and misinformation on its News Feed. (Read more about our partnership with Facebook.) Foxworthy is famous for his \"You might be a redneck if 2026 jokes , but searching online we couldn't find any credible evidence that he penned this list that touches on abortion, Muslims and federal debt. Rather, we found users on web forums crediting someone named Fritz Edmunds with the list. Snopes, which fact-checked this claim back in 2013, also noted that the original compiler of this appears to be Fritz Edmunds, who posted it to his 2018 Politically True blog back in Feb. 3, 2013 (albeit with a disclaimer noting that 2018some of the ideas were from an email that did not contain a copyright). The link to that post is now dead but the url includes the words channeling Jeff Foxworthy. \\

            \arrayrulecolor{black}\midrule
            \multicolumn{1}{c}{\textbf{Initial Explanation}} \\
            \arrayrulecolor{black!30}\midrule
            The claim is labeled as false. \textcolor{blue}{The claim suggests that comedian Jeff Foxworthy wrote a list of examples explaining how "you might live in a nation that was founded by geniuses but is run by idiots."} However, this claim is inaccurate and misleading. \textcolor{red}{There is no credible evidence to suggest that Jeff Foxworthy wrote the specific list mentioned in the claim.} … \\

            \arrayrulecolor{black}\midrule
            \multicolumn{1}{c}{\textbf{Feedback 1}  $F_{0, 1}$} \\
            \arrayrulecolor{black!30}\midrule
            … Extrinsic Event Error: "\textcolor{red}{There is no credible evidence to suggest that Jeff Foxworthy wrote the specific list mentioned in the claim.}"
            Feedback: The statement is too definitive in asserting that there is no credible evidence. It would be more accurate to say that there is \textcolor[rgb]{0.8,0.8,0.0}{currently no credible evidence available.} … \\

            \arrayrulecolor{black}\midrule
            \multicolumn{1}{c}{\textbf{Feedback 2}  $F_{0, 2}$} \\
            \arrayrulecolor{black!30}\midrule
            … Intrinsic Entity-Related Error: The generated explanation incorrectly states, "\textcolor{blue}{The claim suggests that comedian Jeff Foxworthy wrote a list of examples explaining how 'you might live in a nation that was founded by geniuses but is run by idiots.'}" This is incorrect as the claim does not suggest that Jeff Foxworthy wrote the list, but rather falsely attributes it to him. …
            Revise the sentence to accurately state, "The claim \textcolor[rgb]{0.8,0.8,0.0}{falsely} attributes the list to Jeff Foxworthy, suggesting that he wrote it, which is not supported by any credible evidence.” \\

            \arrayrulecolor{black}\midrule
            \multicolumn{1}{c}{\textbf{Judge}  $J$} \\
            \arrayrulecolor{black!30}\midrule
            False \\

            \arrayrulecolor{black}\midrule
            \multicolumn{1}{c}{\textbf{Feedback 1}  $F_{1, 1}$} \\
            \arrayrulecolor{black!30}\midrule
            1. Extrinsic Event Error: "\textcolor{red}{There is no credible evidence to suggest that Jeff Foxworthy wrote the specific list mentioned in the claim.}” … \\ 
            
            2. Intrinsic Entity-Related Error: "\textcolor{blue}{The claim suggests that comedian Jeff Foxworthy wrote a list of examples explaining how 'you might live in a nation that was founded by geniuses but is run by idiots.'}" This assertion is inaccurate because the claim doesn't imply that Jeff Foxworthy wrote the list; instead, it falsely assigns authorship to him. \\

            \arrayrulecolor{black}\midrule
            \multicolumn{1}{c}{\textbf{Feedback 2}  $F_{1, 2}$} \\
            \arrayrulecolor{black!30}\midrule
            1. Intrinsic Entity-Related Error: The generated explanation incorrectly states, "\textcolor{blue}{The claim suggests that comedian Jeff Foxworthy wrote a list of examples explaining how 'you might live in a nation that was founded by geniuses but is run by idiots.'}”\\ ...
            
            2. Extrinsic Event Error: "\textcolor{red}{There is no credible evidence to suggest that Jeff Foxworthy wrote the specific list mentioned in the claim.}” … \\

            \arrayrulecolor{black}\midrule
            \multicolumn{1}{c}{\textbf{Judge}  $J$} \\
            \arrayrulecolor{black!30}\midrule
            True \\

            \arrayrulecolor{black}\midrule
            \multicolumn{1}{c}{\textbf{Accumulated Feedback}} \\
            \arrayrulecolor{black!30}\midrule
            1. Extrinsic Event Error: \textcolor{red}{There is no credible evidence to suggest that Jeff Foxworthy wrote the specific list mentioned in the claim.} The assertion is overly definitive in claiming the absence of credible evidence. A more accurate statement would be that \textcolor[rgb]{0.8,0.8,0.0}{currently, no credible evidence is available}.\\
            2. Intrinsic Entity-Related Error: \textcolor{blue}{The claim suggests that comedian Jeff Foxworthy wrote a list of examples explaining how 'you might live in a nation that was founded by geniuses but is run by idiots.'} The assertion inaccurately suggests Jeff Foxworthy as the author of the list, rather than implying that the claim \textcolor[rgb]{0.8,0.8,0.0}{falsely} attributes authorship to him.\\

            \arrayrulecolor{black}\midrule
            \multicolumn{1}{c}{\textbf{Revised Explanation}  $E^*$} \\
            \arrayrulecolor{black!30}\midrule
            The claim \textcolor[rgb]{0.8,0.8,0.0}{falsely} suggests that Jeff Foxworthy wrote a list of examples explaining how 'you might live in a nation that was founded by geniuses but is run by idiots.' However, \textcolor[rgb]{0.8,0.8,0.0}{as of now}, there is no credible evidence to suggest that Jeff Foxworthy wrote the specific list mentioned in the claim. … \\

            \arrayrulecolor{black}\bottomrule
        \end{tabular}
    \end{adjustbox}
    \caption{An example output from \modelshortest~. Extrinsic Event Error initially identified from Feedback 1 are marked in \textcolor{red}{red}, while Intrinsic Entity-Related Error initially identified from Feedback 2 are marked in \textcolor{blue}{blue}. Both the suggested revisions from each feedback and the actual revisions made from the initial explanation are highlighted in \textcolor[rgb]{0.8,0.8,0.0}{yellow}.}
    \label{tab:Actual_output_debating}
\end{table*}

\end{document}